\documentclass[12pt]{article}

\usepackage{graphicx}
\usepackage{amssymb}
\usepackage{amsthm}
\usepackage{booktabs}
\usepackage{rotating}
\usepackage{multirow}
\usepackage{eurosym}
\usepackage{amsfonts}
\usepackage{amsmath}
\usepackage{umoline}
\usepackage{caption}
\usepackage{subcaption}
\usepackage{color}
\usepackage{url}
\usepackage{authblk}
\usepackage[english]{babel}

\usepackage{pbox}
\usepackage{footnote}
\usepackage{multirow}
\usepackage{changepage}
\usepackage{tablefootnote}

\usepackage[margin = 2cm]{geometry}

\usepackage{hyperref}

\usepackage[natbibapa]{apacite}
 
\begin{document}

\title{The value of text for small business default prediction: A deep learning approach \footnote{\scriptsize NOTICE: This is a preprint of a published work. Changes resulting from the publishing process, such as editing, corrections, structural formatting, and other quality control mechanisms may not be reflected in this version of the document. Please cite this work as follows: Stevenson, M., Mues, C., \& Bravo, C. (2021). The value of text for small business default prediction: A deep learning approach. European Journal of Operational Research 295 (2): 758-771. DOI: \protect\url{https://doi.org/10.1016/j.ejor.2021.03.008}.}}

\author[1,2]{Matthew Stevenson}
\author[1,2]{Christophe Mues}
\author[3]{Cristi\'{a}n Bravo}

\affil[1]{Department of Decision Analytics and Risk, University of Southampton, University Road, SO17 1BJ, United Kingdom.}

\affil[2]{Centre for Operational Research, Management Sciences and Informations Systems (CORMSIS), University of Southampton, University Road, SO17 1BJ, United Kingdom.}

\affil[3]{Department of Statistical and Actuarial Sciences, The University of Western Ontario,1151 Richmond Street, London, Ontario, N6A 3K7, Canada.}

\date{}

\maketitle

\begin{abstract}
Compared to consumer lending, Micro, Small and Medium Enterprise (mSME) credit risk modelling is particularly challenging, as, often, the same sources of information are not available. Therefore, it is standard policy for a loan officer to provide a textual loan assessment to mitigate limited data availability. In turn, this statement is analysed by a credit expert alongside any available standard credit data. In our paper, we exploit recent advances from the field of Deep Learning and Natural Language Processing (NLP), including the BERT (Bidirectional Encoder Representations from Transformers) model, to extract information from 60000 textual assessments provided by a lender. We consider the performance in terms of the AUC (Area Under the receiver operating characteristic Curve) and Brier Score metrics and find that the text alone is surprisingly effective for predicting default. However, when combined with traditional data, it yields no additional predictive capability, with performance dependent on the text's length. Our proposed deep learning model does, however, appear to be robust to the quality of the text and therefore suitable for partly automating the mSME lending process. We also demonstrate how the content of loan assessments influences performance, leading us to a series of recommendations on a new strategy for collecting future mSME loan assessments.
\end{abstract}


\section{Introduction}\label{intro}
Micro, small and medium-sized firms, grouped under the acronym mSME, face many barriers in the process of acquiring credit when compared to their larger and more established counterparts. These barriers have been especially present since the 2008 financial crash when banks became more risk-averse, opting to restrict the availability of credit to larger organisations with greater available credit history \citep{cowling2012small}. Such challenges were echoed once more in 2020 with the outbreak of the COVID-19 virus, impacting small businesses on an international scale. While the long-term impact on mSMEs is unknown, early analysis has highlighted such organisations' financial fragility and the critical role that access to funding will play \citep{bartik2020small}. 

In most personal and established business credit lending instances, there is a wealth of information, including credit history and demographic data, by which a lender can assess credit risk. Such information is often not available for many mSMEs, or at least not to the same extent, presenting a challenge for credit lenders \citep{bravo2013granting}. Information asymmetry exists whereby the mSME knows the business and the context in which it operates; however, the lender, without this knowledge and expertise, cannot confidently assign risk to a loan \citep{lee2015access}. Therefore, it is common in mSME lending for a loan officer to engage with mSMEs to document a company's context and the nature of the loan requirement. The limited structured data relating to the firm's financial performance and economic context can be transformed into a stand-alone credit score, commonly derived using generalised linear models \citep{calabrese2013modelling,calabrese2014generalized}. The credit score supplemented with additional information, including the text, is then reviewed by a credit expert, leading to an accept/reject decision. While this process has been effective \citep{netzer2018words}, collecting and processing this additional information is time-consuming, and it requires expert knowledge to make and interpret the assessment. The result is a disproportionate cost per loan ratio compared to large business lending (higher value) and personal lending (highly automated). Though costly, previous research has identified this process as a necessity in micro-lending as credit scoring models alone are typically weak predictors of loan default \citep{van2012credit}. In addition to resulting in a high cost per loan, the process is inherently reliant on sub-optimal heuristics to assess the input sources' weighting. While this expert knowledge brings value, in the current setting, it is almost impossible to evaluate whether the correct (relative) weights are given to the input sources or whether this knowledge is founded on outdated, incorrect or even biased decisions. In this paper, we put forward an attractive alternative based on machine learning; i.e.\@ , we build a state-of-the-art deep learning model that simplifies the latter stage of the process. We argue that our approach improves the status quo in three ways:
\begin{itemize}
	\item Our proposed method can speed up the interpretation process by extracting an automated risk score from the text that we show to be predictive.
	\item It is capable of combining these two separate sources of information (i.e.\@ text and structured data) and suggesting a single score, and a corresponding accept/reject decision.
	\item By adding an extra explanation layer in our approach, insights are produced where both sources of information complement or contradict each other, which can be fed back to improving both the lender's structured data and text collection practices.
\end{itemize}

Several studies have sought to leverage unstructured text data for credit scoring; however, these have been restricted to Peer-to-Peer (P2P) lending and seemingly have not yet exploited the recent advances from Deep Learning. There have however been several examples of Deep Learning for financial applications, including fraud detection \citep{rushin2017horse,wang2018leveraging}, mortgage risk \citep{kvamme2018predicting,sirignano2016deep}, bankruptcy prediction \citep{chen2018role,dorfleitner2016description,jiang2018loan,netzer2018words}, churn prediction \citep{spanoudes2017deep} and financial market prediction \citep{fischer2018deep}. More broadly, we also observe that many studies that consider the predictive capacity of text tend to consider it in isolation. While yielding sometimes impressive results, they give little consideration as to how textual data might be integrated with existing structured data. To the best of our knowledge, this paper is the first to apply the recent advances in deep learning to the problem of mSME loan default prediction, integrating both text and structured data. Specifically, we contribute to the literature by applying one of the current state-of-the-art language models, BERT (Bidirectional Encoder Representations from Transformers), to a real-life mSME loan dataset. We evaluate the predictive performance of the proposed approach using the Area Under the receiver operating characteristic Curve (AUC) and Brier Score metrics for two subsets of customers: new and existing. In addition to reviewing performance, we also provide unique insights into how the text data influences credit default predictions. We do so by assessing the impact of the text on variable importance and predicted probabilities; additionally, we evaluate how the text's content may lead to better or worse predictions.

Our research questions thus are:
\begin{enumerate}
	\item Are the textual loan officer assessments predictive of loan default and if so, does Deep Learning provide an improved result over traditional text-mining and machine learning approaches?
	\item Does the inclusion of the text alongside traditional structured sources of data lead to an enhanced result over using the structured data alone? If so, are Deep Learning approaches better suited for integrating the combined text and structured inputs to produce an automated risk score?	
	\item To what extent does the inclusion of text as a model input impact how the model makes its predictions and the resulting risk ranking?
	\item In what instances does the text's content contribute positively to predictive capacity, and what insights can be gained from this for future loan assessment collections?
\end{enumerate}

The remainder of the paper is structured as follows. First, in section \ref{literatureReview} we summarise the recent advancements in Deep Learning and NLP. Then in section \ref{heading:descofdata}, we present an overview of the dataset, including a description of the available text statements. Following this, in section \ref{heading:models}, a description of the modelling approach is provided, before presenting the performance results in section \ref{heading:results}. We then look beyond the performance results and provide an analysis into how the presence of text influences the behaviour of a model (section \ref{heading:expimpactoftext}) by reviewing feature importance (section \ref{subheading:featureimportance}), the impact of the text on the predicted probabilities (section \ref{subheading:ImpactonProbabilities }) and how the content of the text influences the predictive capacity of the model (section \ref{subheading:importantwords}). Finally, we summarise our findings and potential implications in sections \ref{heading:summaryandimplications} and \ref{heading:conclusion}.
 
 \section{Literature Review}\label{literatureReview}
In predictive analytics, machine learning attempts to output a prediction for a task given a set of inputs. Deep Learning is a collection of techniques that allow representations to be learned from complex structures. In the context of predictive analytics, it can be used to provide a prediction from raw data \citep{lecun2015deep}. While in traditional machine learning, when working with unstructured data, features are hand-crafted from the data by experts or by indirect statistical transformations, with Deep Learning, the data can be processed in almost raw form. In the past decade, Deep Learning techniques have been applied to unstructured data sources such as audio, images and text, achieving state-of-the-art results across various predictive tasks. Furthermore, although deep learning is typically associated with performance breakthroughs on unstructured data, in some instances, the associated tools and techniques have led to improvement on structured data sources. For example, embedding techniques which had been successfully applied to text data have also shown to be effective for datasets containing categorical features with high cardinality, leading to improved performance and generalisability \citep{guo2016entity}. 

The Natural Language Processing (NLP) domain has seen significant improvements in predictive capabilities since the introduction of Deep Learning techniques. Before the introduction of Deep Learning, pre-defined or statistical-based rules were required to extract features from text data. Dictionary methods based on expert knowledge are perhaps the simplest form of feature extraction, whereby general or domain-specific words are extracted from the text and treated as feature inputs \citep{al2017approaches}. Alternatively, approaches such as bag-of-words \citep{tsai2017risk}, Latent Dirichlet Allocation (LDA) \citep{blei2003latent} and Latent Semantic Analysis (LSA) \citep{landauer1998introduction} derive features based on statistical rules and do not require human labelling. Such approaches were and remain popular due to their relative simplicity and interpretability. Indeed, these approaches have been applied to a range of NLP tasks, e.g.\@ predicting financial risk from financial reports \citep{tsai2017risk}, classifying fake news \citep{zhang20191036}, interpreting the risk culture of banks \citep{agarwal2019learning}. In contrast to earlier rule-based or statistics-based approaches, Deep Learning models can effectively extract features from the raw text with little prior pre-processing requirements. One of the first advances here is the introduction of word embeddings which allows words to be efficiently represented by a vector of numbers that can be passed to a machine learning model \citep{turian2010word}. 

Two types of models, in particular, have shown state-of-the-art results: the Convolutional Neural Network (CNN) and the Recurrent Neural Network (RNN) \citep{zhang2015character}. CNN models allow for representations to be learned from n-grams (groups of neighbouring words) derived from a fixed-sized text sequence. RNNs, on the other hand, are sequence-to-sequence models that do not require a fixed-size input. RNNs have a memory state that processes a single word or word piece at a time and can retain and forget representations as necessary; in the context of text classification, it is common that the states of the model across a text statement are aggregated. The aggregated layers of the RNN model provide a context vector for the entire sequence, which is used in turn to produce a classification. Both the CNN and RNN models have demonstrated improved results over earlier machine learning approaches and enabled novel business applications.

More recently, NLP has also seen the introduction of two techniques that have enabled further performance improvements: Transfer Learning and Attention models. Transfer learning is the concept that a model trained on one task can be fine-tuned to another similar task \citep{howard2018fine}. While commonplace in image processing previously, since 2017, NLP has seen the introduction of models that utilise transfer learning, including ELMO \citep{peters2018deep}, Ulm-Fit \citep{howard2018fine}, BERT \citep{devlin2018bert} and XLNET \citep{yang2019xlnet}, which are now the state of the art across many NLP benchmarking tasks. Attention and Self-Attention are mechanisms used in Deep Learning architectures that allow these models to pay attention to specific words in a sentence, resolving some of the long-term dependency issues that historically exist with CNN and RNN models \citep{bahdanau2014neural,vaswani2017attention}. Furthermore, attention has been effectively utilised within Transformer Models, a flavour of encoder-decoder models that allow for complex models to be efficiently parallelised \citep{vaswani2017attention}. Parallelisation allows for quicker training times and, in some instances, provides a boost in predictive performance.

There are several examples of how NLP has been applied to predict credit risk from user-generated text in consumer P2P lending. Interestingly, these studies have yielded mixed results in terms of the predictive capacity of text. \citet{dorfleitner2016description}, for example, derive features using text from the P2P platforms Smava and Auxmoney and find that aspects of the text such as the number of spelling mistakes, text length and emotive language are predictive of funding likelihood but not of default. Similarly, \citet{chen2018role} explore the predictiveness of the punctuation used in user-generated text from a large P2P lender, Renrendai, and do not find a significant relationship with loan default. Conversely, \citet{netzer2018words} use text from the P2P platform Prosper and find the text to be predictive of loan default, with a modest but statistically significant uplift of 2.64\% in AUC when using both text and structured features. \citet{jiang2018loan} also derive topics from P2P models using LDA and have shown their best performing Random Forest model provides a 2.3\% uplift in AUC when using both soft (text) and hard (structured) features. 

Deep Learning NLP approaches have also been applied to bankruptcy prediction, with some success. \citet{mai2019deep}, for example, apply Deep Learning techniques to a bankruptcy database of 11,827 U.S public companies using text from disclosure statements, which typically include details such as a description of the business activities, financial performance and organisational structure. When the text is used alongside accounting and market-based data, Deep Learning was found to improve predictive accuracy. Similarly, \citet{matin2019predicting} predict bankruptcy using text segments from annual reports for Danish non-financial and non-holding private limited and stock-based firms and find that the text augmented with structured features leads to improved prediction. These studies are significant as they demonstrate that Deep Learning can improve predictions of bankruptcy when using structured and unstructured features. Bankruptcy prediction is, however, inherently different from credit application scoring. Firstly, credit application scoring is an activity driven by a lender to inform loan acceptance policies, while bankruptcy is a legal outcome for an organisation that cannot meet its financial obligations. Furthermore, in terms of definition, an organisation can default on a loan without falling into bankruptcy. Conversely, though to a lesser extent, an organisation can fall into bankruptcy and fulfil some or all its credit obligations to a lender. Importantly, we also note that these studies relate to larger firms and utilise longer text sources.

To conclude, the exploration of text for predicting default (or default-like) outcomes has tended to be restricted to P2P lending and bankruptcy prediction. In research in the P2P field, utilising user-generated text has yielded mixed results. These studies are also yet to utilise Deep Learning techniques, which have shown improved results for structured and unstructured data. Such techniques have been applied for bankruptcy prediction for publicly listed companies, using typically quite large texts. These studies however are yet to utilise the latest state-of-the-art NLP models. Our work thus makes the following contributions. To the best of our knowledge, we are the first to review the predictive performance of text alongside structured information in the context of mSME credit scoring. Furthermore, our dataset is novel, not just because of the nature of the obligors it contains but also the way the text is generated, i.e.\@ by expert loan officers rather than the credit applicant. In our research, we also use the latest generation of transformer text models, in the form of BERT (Bidirectional Encoder Representations from Transformers), a model capable of exploiting deep semantic meaning from the loan assessments. We look beyond the pure predictive task and explore how text, when introduced in the model, changes the nature of the output prediction. We also explore what content of the loan assessments can lead to improved performance and how this might influence the future data collection for the mSME lender.

\section{Description of the data} \label{heading:descofdata}
\vspace{3mm}
\subsection{Overview}
The dataset belongs to a specialised Micro and SME lender based in a medium-high income country in South America. This lender provides loans nationwide across various industries, typically of length 6 to 24 months in duration. The dataset contains 60000 valid records complete with 43 features, in addition to the one text feature provided by a credit officer. The loan dataset extends over the period 2008 to 2015, in which the country of operation saw a period of sustained growth. Default is the dependent variable, defined as 90 days past due over the 18 month period of observation, in alignment with the Basel Accords definition \citep{siddiqi2017intelligent}.

\subsection{Standard features}
The raw dataset contains 43 standard credit scoring features representing demographic information (e.g.\@ age, schooling, civil status, the region of the country), business information (e.g.\@ age of business, economic area, sales) and information relating to the loan (e.g.\@ loan value, any previous credits, credit type). Before modelling, we apply a Random Forest with Recursive Feature Elimination and Cross-Validation (RFECV) to select an appropriate subset of features \citep{granitto2006recursive}. This step is required to remove redundancy in the variables and eliminate correlated variables that impede the feature importance analysis. The RFECV approach trains a model and recursively drops the least important variable until only a single variable remains; this process is repeated over ten folds, and the model is trimmed to include the minimum number of variables without impeding the cross-validated AUC performance. We find 26 structured variables to be the optimal number of features providing the highest AUC score. 

We apply Min/Max scaling to the continuous variables and introduce label encoding for the categorical variables.

\subsection{Text description}

It is the role of a loan officer from the credit lender, who usually visits the applicant, to gather contextual information. This summary typically includes a short description of the applicant's business context, the requirement of the loan, and, if appropriate, the client's payment history with the lender. 

\begin{figure}[!htb]
\centering
\scalebox{0.95}{
  \includegraphics[width=1\linewidth]{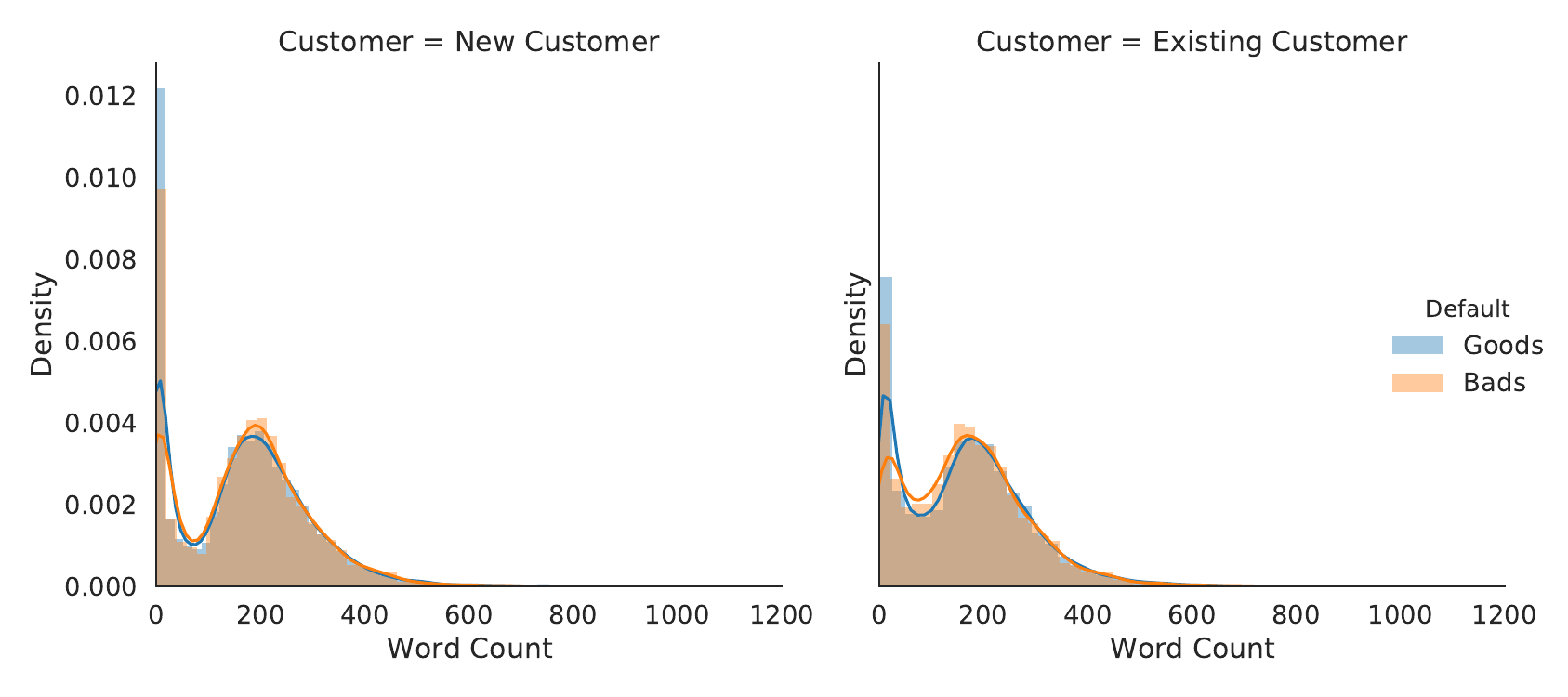}}
  \caption{Loan assessment word count}
  \label{fig:wrdcnt1}
\end{figure}

Figure \ref{fig:wrdcnt1} presents the word count for the Goods (non-defaulters) and Bads (defaulters) across two customer types --- New and Existing. The length of the text assessments assumes a bimodal distribution for both customer types, whereby there is a group of loans with perhaps just a sentence and a group of more extensive descriptions with a mean word count of 200 words extending up to 1200 words. Although the text length distributions for Goods and Bads are similar, we see that typically the Goods have a higher frequency of lower word counts across both customer types. This intuitively suggests that the loan officers generate longer and more detailed texts for loans that are deemed riskier.

Below we include some typical extracts from the text assessments (translated from the original text).

\textbf{Extract 1}:
``\textit{The partner is a small informal farmer, who works and lives on their land. They are requesting the loan for the purchase of agricultural inputs, mainly fertilisers for the future sowing of ?? hectares of land}"

\textbf{Extract 2}:
``\textit{The firm's sales are around \$???? per month from the sale of bricks, aggregates and firewood. It sells between ??? and ??? bricks monthly. The company's operating costs are derived from freight payments of its materials, which are on average \$??? per month. They are requesting \$???? to buy materials for the preparation of the bricks and some arid materials}"

\textbf{Extract 3}:
``\textit{The client has an excellent business background, the approval for an amount of \$???? for an ?? month term is recommended. The client intends to invest in the purchase of alternative cosmetics and perfumes which they intend to market to clients during the holiday season"}

Note that the text is fully anonymised, masking all entity names, dates, locations and quantitative values. We also apply basic cleaning operations to the text to remove common errors, e.g.\@ redundant white spacing. 

Two types of pre-processing are applied before fitting the three types of models, i.e.\@ Logistic Regression (LR), Random Forests (RF) and Deep Learning (DL). 

First, for the DL model, which utilises Google's pre-trained BERT model \citep{devlin2018bert}, the word pieces are mapped to a pre-existing vocabulary in the pre-trained model. The BERT model requires a maximum token length to be defined; we select 512 words per statement as this is close to the 99th percentile of 524 words but more computationally efficient. Additionally, we choose a higher percentile threshold for the documents to allow some redundancy as the BERT model further splits some words to part-of-word segments. The full details of the subsequent pre-processing for the BERT model are detailed later in this paper. 

Second, for the LR and RF models, unlike with the DL approach where the text representation learning forms an active part of the model training, such an approach is not directly compatible. Instead, a representation of the text statements must be pre-calculated using a statistical method to produce a 1-dimensional vector per text statement. To produce the vector representations, we apply Latent Semantic Analysis (LSA) \citep{landauer1998introduction}, a two-stage process commonly applied in NLP applications that extract representations based on pre-defined statistical rules. First, a Term Frequency-Inverse Document Frequency (TF-IDF) matrix is derived of length \textit{no. statements} and width \textit{vocab size}, providing a vector for each text statement reflecting how important each word is in a text statement, relative to the collection of statements \citep{ramos2003using}. At this step, we remove unnecessary stop words and additionally screen words that occur in more than 10\% or less than 5\% of the texts. The TF-IDF matrix is large as all words in the \textit{vocab size} axis are represented for all statements. This generates redundancy as there are many correlated and unused words for each statement. Therefore, the second stage of LSA is to apply dimensionality reduction. We use a truncated singular-value decomposition (SVD) \citep{husbands2001use} to produce `concepts' that explain 94\% of the total variance of the TF-IDF matrix. The result is a vector of length 250 for each text statement.

\section{Models}\label{heading:models}
\vspace{3mm}
\subsection{Model descriptions}
We seek to assess both the predictive power of the text and of the modelling techniques. Accordingly, we consider three types of model (i.e.\@ LR, RF and DL) and three subsets of data; text (the text alone), structured (standard credit scoring features alone) and combined (both the structured and text inputs). The three models and three data subsets produce the nine models listed in Table \ref{InputTable}.

The LR and RF models are used as baseline models against which to compare the DL models' performance. For the LR, the simplest model due to its linearity assumption, we use an Elastic Net implementation, that uses L1(Lasso)/L2(Ridge) regularisation \citep{friedman2010regularization}. Across all LR models, we optimise the \textit{L1/L2 ratio} parameter using a grid search with 10 folds of the training data. We also apply a non-linear RF classifier, which should theoretically perform better than the LR due to its capability to capture non-linear features known to exist within text data \citep{kim2014sentiment}. We similarly use a Random Grid Search with 10 fold cross-validation to optimise the parameters \textit{max depth} and \textit{max features}. The number of trees (\textit{ntree} parameter) is set to 10000 as it has been demonstrated that little performance is gained by adding additional trees beyond this threshold for medium-sized datasets such as this one \citep{probst2017tune}. To produce the LR and RF models for the combined data (i.e.\@ structured+text), we concatenate both sets of inputs to create a single vector input per record.

\begin{table}[!htb]
\caption{Inputs by Model and Data Subset}
\centering
\scalebox{0.8}{
\begin{tabular}{@{}llll@{}}
\toprule
           & \multicolumn{1}{c}{LR}           & \multicolumn{1}{c}{RF}            & \multicolumn{1}{c}{DL}            \\ \midrule
Structured & \multicolumn{1}{c}{13 Continuous} & \multicolumn{1}{c}{13 Continuous} & \multicolumn{1}{c}{13 Continuous} \\
           & 13 Categories                     & 13 Categories                     & 13 Categories                     \\
           &                                   &                                   &                                   \\
Text       & \multicolumn{1}{l}{250 Concepts}  & \multicolumn{1}{l}{250 Concepts}  & \multicolumn{1}{l}{512 Tokens}    \\
           &                                   &                                   &                                   \\
Combined   & \multicolumn{1}{c}{13 Continuous} & \multicolumn{1}{c}{13 Continuous} & \multicolumn{1}{c}{13 Continuous} \\
           & 13 Categories                     & 13 Categories                     & 13 Categories                     \\
           & 250 Concepts                      & 250 Concepts                      & 512 Tokens                        \\ \bottomrule
\end{tabular}}
\label{InputTable}
\end{table}

In addition to these benchmark models, three deep learning models are trained --- one on the structured data only, another on the (pre-processed) text data, and a third on the combined data. However, there are notable differences between the DL architectures chosen for the structured and the text data.

Firstly, at the core of the structured DL model is a simple Multi-layer Perceptron (MLP); however, in the MLP, we implement the relatively new concept of categorical embeddings \citep{guo2016entity}. Categorical embeddings were inspired by developments in the Deep Learning and NLP community that have allowed words to be represented as a numerical vector, by performing language tasks on a text corpus. For example, the well-known Word2Vec model implements a neural network to predict a word, given neighbouring words in a sentence. The single hidden layer of the Word2Vec neural network model is subsequently used to represent each word within a given vocabulary \citep{pennington2014glove}. The result is a multi-dimensional space in which words with similar semantic meaning are mapped closely together and can be used for a downstream NLP task, such as sentiment classification. The benefit of such an approach is that it allows meaningful representations of a word to be extracted that could not be achieved with one-hot encoding due to the naturally high cardinality of words.

\begin{figure}[!htb]
\centering
\scalebox{0.7}{
  \includegraphics[width=1\linewidth]{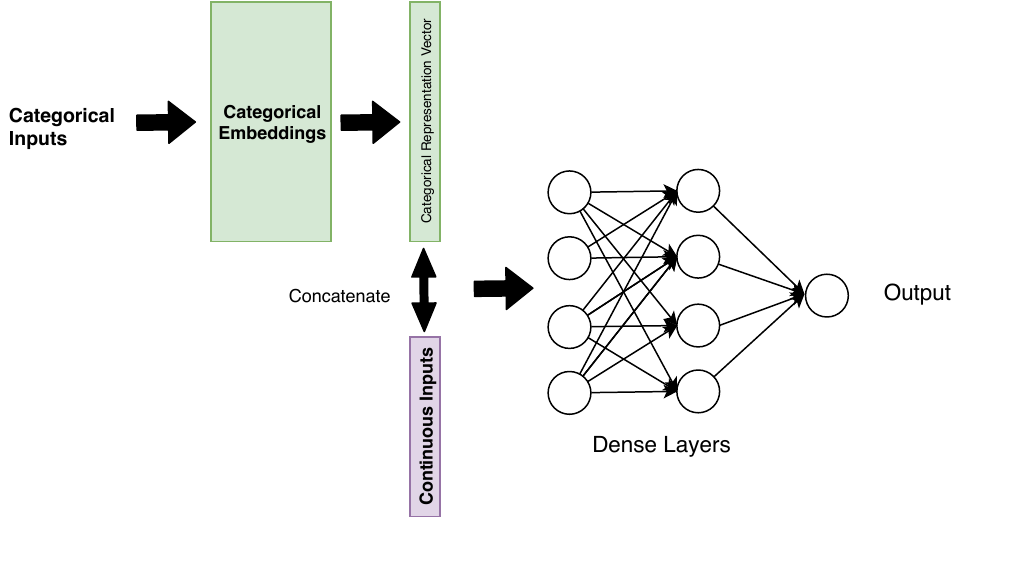}}
  \caption{Structured Model Architecture}
  \label{fig:StrucNN}
\end{figure}

Similarly, we implement categorical entity embeddings, which may help categorical data with high cardinalities such as the residing town or branch district. There are some fundamental differences between word embeddings and categorical embeddings, however. Firstly, we note that the embedding layer training is not unstructured as the features are mapped in the context of $P$(Default). We also note that unlike word embeddings, which have a shared vocabulary and embedding space, in our model, each vocabulary and embedding is unique to the particular categorical feature. The size of the embedding space is also unique to each categorical feature. For word embeddings, the vector size can be many times smaller than that of the vocabulary size. For example, \citet{pennington2014glove} found diminishing returns beyond a vector length of just 300 for their GloVe embeddings, using a vocabulary size of 40000 words. We set the length of each embedding to $\lfloor{\frac{n}{2}}\rceil$, where $n$ is the number of unique categories in the feature. This rule, therefore, seeks to allow redundancy, ensuring that the categorical embeddings are not under-sized. Each of the categorical feature embeddings is concatenated together with the continuous and binary features before being passed to the model's subsequent dense layer (see Figure \ref{fig:StrucNN}).

\begin{figure}[!htb]
\centering
\scalebox{0.7}{
  \includegraphics[width=1\linewidth]{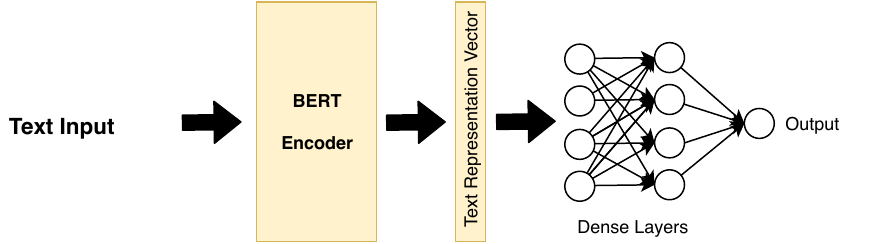}}
  \caption{Text-Only Model Architecture}
  \label{fig:bert1}
\end{figure}

Secondly, for the text-only DL model, we use BERT, a state-of-the-art language model produced by researchers at Google AI Language \citep{devlin2018bert}. This model has achieved promising results across a range of common NLP tasks and builds upon the concept of transfer learning. Transfer learning uses a pre-trained model applied to similar or ancillary tasks and applies a strategy to extract information from the pre-trained model, either by using a model's fixed outputs or by fine-tuning the original model to a new task \citep{devlin2018bert}. The BERT model is trained on two ancillary tasks: 1) predicting a masked word in a sentence, and, 2) next sentence prediction. As our text data is in Spanish, we apply the \textit{BERT-Base: Multilingual Cased} implementation of the model which has been trained using word-pieces from the Spanish vocabulary. This model has been shown to provide surprising adaptability across both a range of tasks and languages, even without further fine-tuning (zero-shot learning) \citep{pires2019multilingual}. From the base model, we fine-tune the final 10 layers of the model to further adapt to the language (Spanish) and context (Credit Scoring). We then extract a pooled representation producing a fixed-length vector of 768, which is subsequently connected to an MLP layer classifier (see Figure \ref{fig:bert1}).

Thirdly, the Deep Learning model for the combined data extends the model for the structured data with a fixed-length representation vector from the text-only BERT model. We do so by first independently training the structured and text models, which become the combined model's tails. In the penultimate dense layer, each respective model is merged, and a new head of the model is attached consisting of 712 dense units. The combined model is then passed through two further training rounds -- initially with the tail weights of the model frozen so that learning occurs only in the dense head. Further to this, the entirety of the model is trained together. A higher learning rate is initially applied to the new untrained dense head (1e-2), before using a lower learning rate to fine-tune the full model (5e-6). These approaches are proposed by \citet{howard2018fine} in the context of language model transfer-learning and are referred to as ``Gradual unfreezing" and ``Discriminative [learning rate] fine-tuning". Conceptually, in the context of language transfer learning, the approaches seek to minimise ``forgetting" between the native task and transfer application by differentiating how parts of the model are trained, preserving general information in the earlier layers of the model while using a more aggressive regime on the latter layers. In our combined model, we seek to achieve the same result, preserving as much information in the independently trained tails while allowing information to be integrated in a controlled manner.

\begin{figure}[!htb]
\centering
\scalebox{0.5}{
  \includegraphics[width=1\linewidth]{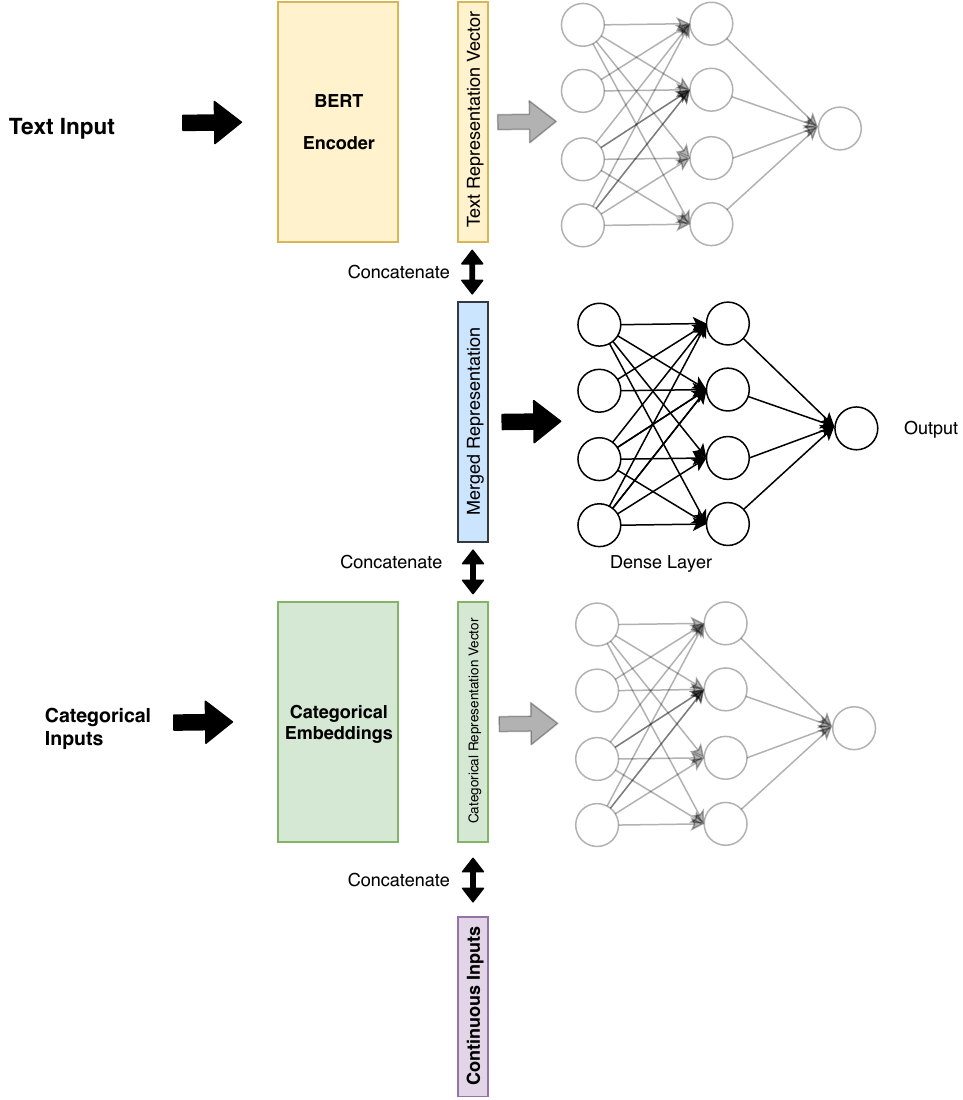}}
  \caption{Combined Model Architecture}
  \label{fig:CombNN}
\end{figure}

An illustration of the combined model architecture is presented in Figure \ref{fig:CombNN}.

\subsection{Model performance assessment}\label{subsection:Model performance assessment}

Performance is assessed using two metrics: AUC (Area Under the receiver operating characteristic Curve) and the Brier Score. 

The AUC metric provides insight into the discriminative capability of a model across a range of cut-off points \citep{baesens2003benchmarking}. The AUC is a particularly insightful metric in practical terms as lenders may choose to alter cut-off points to manage risk tolerance. Furthermore, the metric is not sensitive to class imbalance, as commonly found in credit scoring, where the percentage of defaults is relatively low. 

The second metric we consider is the Brier Score \citep{brier1950verification}. While the AUC metric captures the models' discriminative capacity, the Brier Score is a measure of how well calibrated the output predictions are. The metric is equivalent to the mean squared error but for a binary prediction task. A predicted output closer to the true label (0 Non-Default/ 1 Default) produces a smaller error. As the Brier Score is sensitive to class imbalance, we apply a weighted implementation that balances the two classes. While it is desirable for a model to produce well-calibrated scores, in practice, it is not a requirement for a good classifier as the probability cut-off point can be adjusted accordingly. As such, in our results, we consider the Brier Score as a secondary metric to the AUC. 

We also consider the model's performance across two subsets of the customers: new and existing customers. Existing customers are those who, at the time of their application, had acquired loans with the bank before but were subsequently looking for a new credit line or to refinance an existing loan. New customers, on the other hand, were first-time customers at the lender with no previous account history. This segmentation criterion is common in credit scoring as often the drivers vary (e.g.\@ the importance of past financial performance).

\begin{table}[!htb]
\caption{Data Subsets Description}
\label{table:datasubsettable}
\centering
\scalebox{0.7}{
\begin{tabular}{@{}lccccccc@{}}
\toprule
\multirow{2}{*}{}        & \multicolumn{3}{c}{\multirow{2}{*}{New Customers}}                                            & \multicolumn{1}{l}{} & \multicolumn{3}{c}{\multirow{2}{*}{Existing Customers}}                                       \\
                         & \multicolumn{3}{c}{}                                                                          & \multicolumn{1}{l}{} & \multicolumn{3}{c}{}                                                                          \\ \cmidrule(lr){2-4} \cmidrule(l){6-8} 
\multirow{2}{*}{}        & \multirow{2}{*}{Cases}  & \multirow{2}{*}{\% Default Rate} & \multirow{2}{*}{Mean Word Count} &                      & \multirow{2}{*}{Cases}  & \multirow{2}{*}{\% Default Rate} & \multirow{2}{*}{Mean Word Count} \\
                         &                         &                                  &                                  &                      &                         &                                  &                                  \\ \midrule
\multirow{2}{*}{Train}   & \multirow{2}{*}{17,540} & \multirow{2}{*}{20.60\%}         & \multirow{2}{*}{184}             &                      & \multirow{2}{*}{18,943} & \multirow{2}{*}{10.70\%}         & \multirow{2}{*}{196}             \\
                         &                         &                                  &                                  &                      &                         &                                  &                                  \\
\multirow{2}{*}{HoldOut} & \multirow{2}{*}{4,372}  & \multirow{2}{*}{20.40\%}         & \multirow{2}{*}{185}             &                      & \multirow{2}{*}{4,749}  & \multirow{2}{*}{11.10\%}         & \multirow{2}{*}{197}             \\
                         &                         &                                  &                                  &                      &                         &                                  &                                  \\
\multirow{2}{*}{2008}    & \multirow{2}{*}{1,652}  & \multirow{2}{*}{41.90\%}         & \multirow{2}{*}{236}             &                      & \multirow{2}{*}{14}     & \multirow{2}{*}{42.90\%}         & \multirow{2}{*}{155}             \\
                         &                         &                                  &                                  &                      &                         &                                  &                                  \\
\multirow{2}{*}{2014}    & \multirow{2}{*}{3,585}  & \multirow{2}{*}{20.40\%}         & \multirow{2}{*}{37}              &                      & \multirow{2}{*}{9,095}  & \multirow{2}{*}{11.70\%}         & \multirow{2}{*}{81}              \\
                         &                         &                                  &                                  & \multicolumn{1}{l}{} &                         &                                  &                                  \\ \bottomrule
\end{tabular}}
\end{table}

We use two types of validation. First, the out-of-time samples include the first year (2008) and final year (2014) of the coverage period. We note that typically, out-of-time sampling is conducted using just the end-of-period data. We include the year 2008 (beginning-of-period) as a separate out-of-time sample as it differs structurally from the core data for reasons we subsequently explain. Second, the hold-out sample includes a randomly selected sample of 20\% of the cases between 2009 and 2013, with the remaining 80\% of cases during this period forming the training set. This validation process provides a wide-ranging assessment of our methodology. The out-of-time validation provides insights into different operational and economic conditions, whereas the hold-out sample gives a best-case scenario where the model is applied to unseen, but structurally similar data. Real-life performance would then fall somewhere in between these validation measures.

Table \ref{table:datasubsettable} provides a summary of the key features of each subset. The training and hold-out samples are very closely mirrored with broadly similar default rates and mean word counts. Furthermore, both of these data subsets contain a proportionally similar ratio of new to existing customers. Unlike the hold-out sample, the 2008 and 2014 samples have unique features that differ from the training set. The lender was founded in 2008, and therefore, there are few cases of existing customers; for this reason, we excluded this specific subset from the training sample. The 2008 loan cohort is also characterised as having a higher default rate for new customers --- twice that of the training set --- and, by containing larger and likely more descriptive text information. The 2014 loan cohort, on the other hand, has a default rate in line with the training set but contains considerably shorter texts. The shorter texts in 2014 can be attributed to a change in company procedure for collecting loan assessments.

\section{Results}\label{heading:results}
\subsection{Model performance} \label{subheading:modelperformance}
\begin{figure}[!htb]
\centering
\scalebox{0.8}{
  \includegraphics[width=1\linewidth]{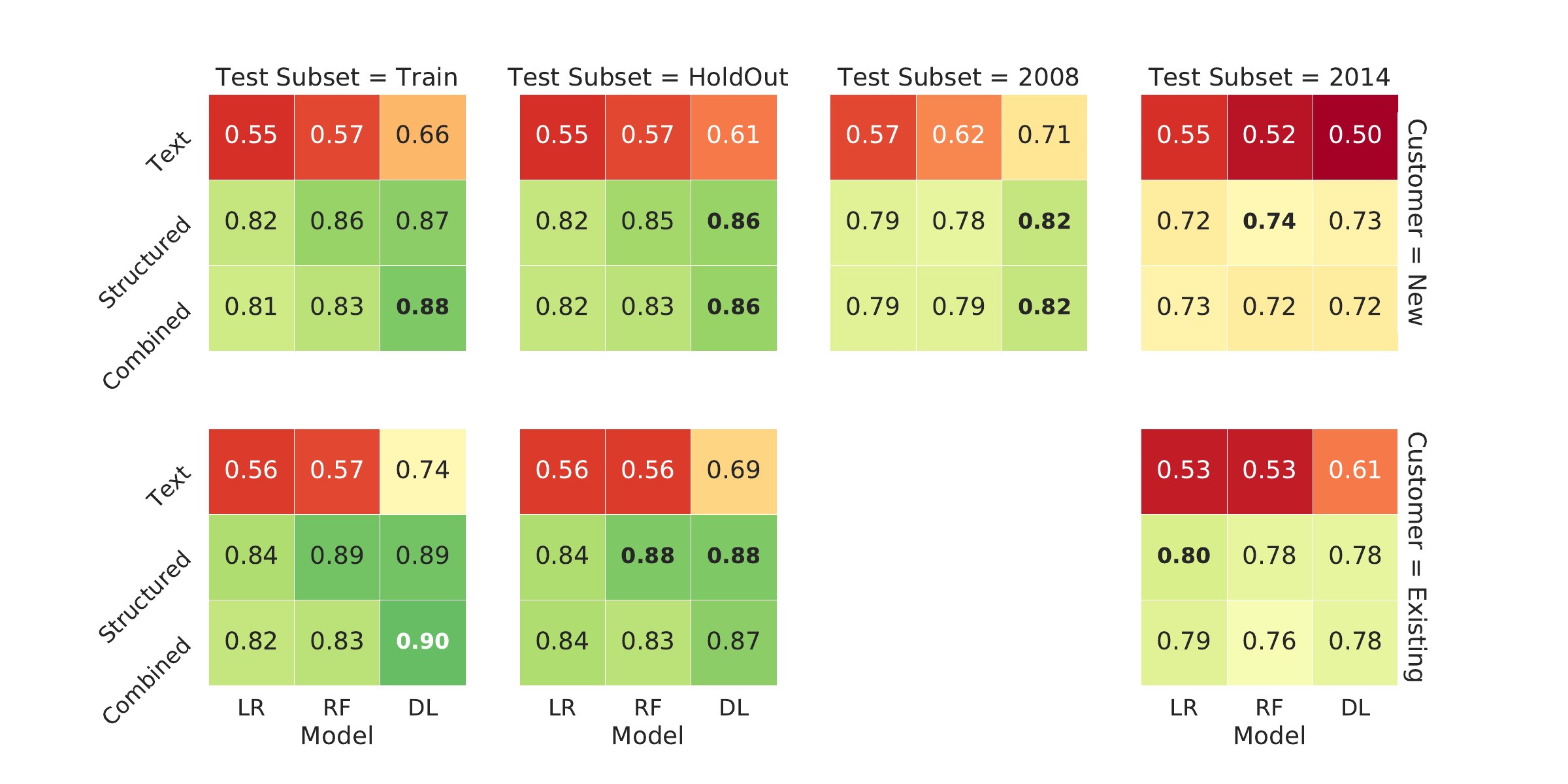}}
  \caption{AUC Model Performance}
  \label{fig:auc_result}
\end{figure}

Figure \ref{fig:auc_result} displays the performance results consisting of seven complete grids. Each grid, in turn, shows nine results. On the y-axis of each grid is the feature subset (Text, Structured, Combined) and on the x-axis is the model (LR, RF, DL). Horizontally, there are four grids, one for each data subset (Train, Hold-out, 2008, 2014) while the top and bottom section of the figure show different grids for new and existing customers, respectively. The best performing model(s) in each grid are presented in bold text format. We note that the grid for existing customers in 2008 is excluded due to the lack of cases, as explained in section \ref{subsection:Model performance assessment}.

Considering the text-only models, we can see that the DL BERT model outperforms the LR and RF models (using an LSA approach), achieving impressive test AUC scores between 0.61 and 0.71. The exception to this is the new customer subset in 2014 where none of the text-only models performed well and the performance of the BERT model was just 0.50 and therefore random. We can likely attribute this difference in performance to the considerably shorter text lengths in 2014 (see Section \ref{subheading:textlength}). For the LR and RF text-only models, there is little performance difference between the new and the existing customers. Conversely, with the DL text model, there is an apparent uplift for the existing customer subset. Given that the text lengths are similar, the difference can likely be attributed to the text's contents. Just as with the structured data, more information is known regarding the existing customer group, which may be reflected in the text. These results suggest that this type of text lends itself better to the more advanced BERT model.

As expected, the structured-only models outperform the text-only models as there is richer information in this source. In most instances, the RF and DL structured models outperform the LR, suggesting non-linearity in the data. Furthermore, the RF and DL models' performance is close; in almost all instances, the difference in AUC is not more than 0.01. The exception to this is for the new customer subset in 2008, where the DL model achieved a higher AUC score (0.82) compared to the LR (0.79) and RF (0.78) models. We attribute this modest performance gain of the DL model to the powerful technique of categorical embeddings. As with the text data, the existing customer group's performance is higher than that of the new customer group where less is known about the applicants.

Despite the promising results of the text-only models, particularly the BERT DL model, when we reviewed the combined model's performance against that of the structured model, the difference is negligible. Indeed, our results suggest that for the RF combined model, the text's introduction can hinder the model, perhaps as a result of the greater complexity. This drop in performance is most significant in the hold-out sample for existing customers, where AUC performance falls from 0.88 to 0.83. The combined LR and DL models, however, appear to be more robust as their performance does not differ by more than 0.01 from the respective structured-only performance across all data subsets. Most importantly, however, the DL combined model is superior compared to both the RF and LR combined models. In almost all instances, the model achieves AUC scores well into the 0.80+ range (up to 0.87) and several percentage points above that of the two benchmark methods. The exception to this is 2014 where, for both subsets of new and existing customers, the performance of all methods drops and, if anything, the LR marginally outperforms the DL model. The drop in performance in 2014 against the hold-out sample can be observed across all models, including the structured-only models. This could be attributed to a potential population shift in the underlying data. Furthermore, it may also indicate that LSA with the LR works reasonably well for shorter texts where more straightforward relationships are likely to exist for the text models.

\begin{figure}[!htb]
\centering
\scalebox{0.9}{
  \includegraphics[width=1\linewidth]{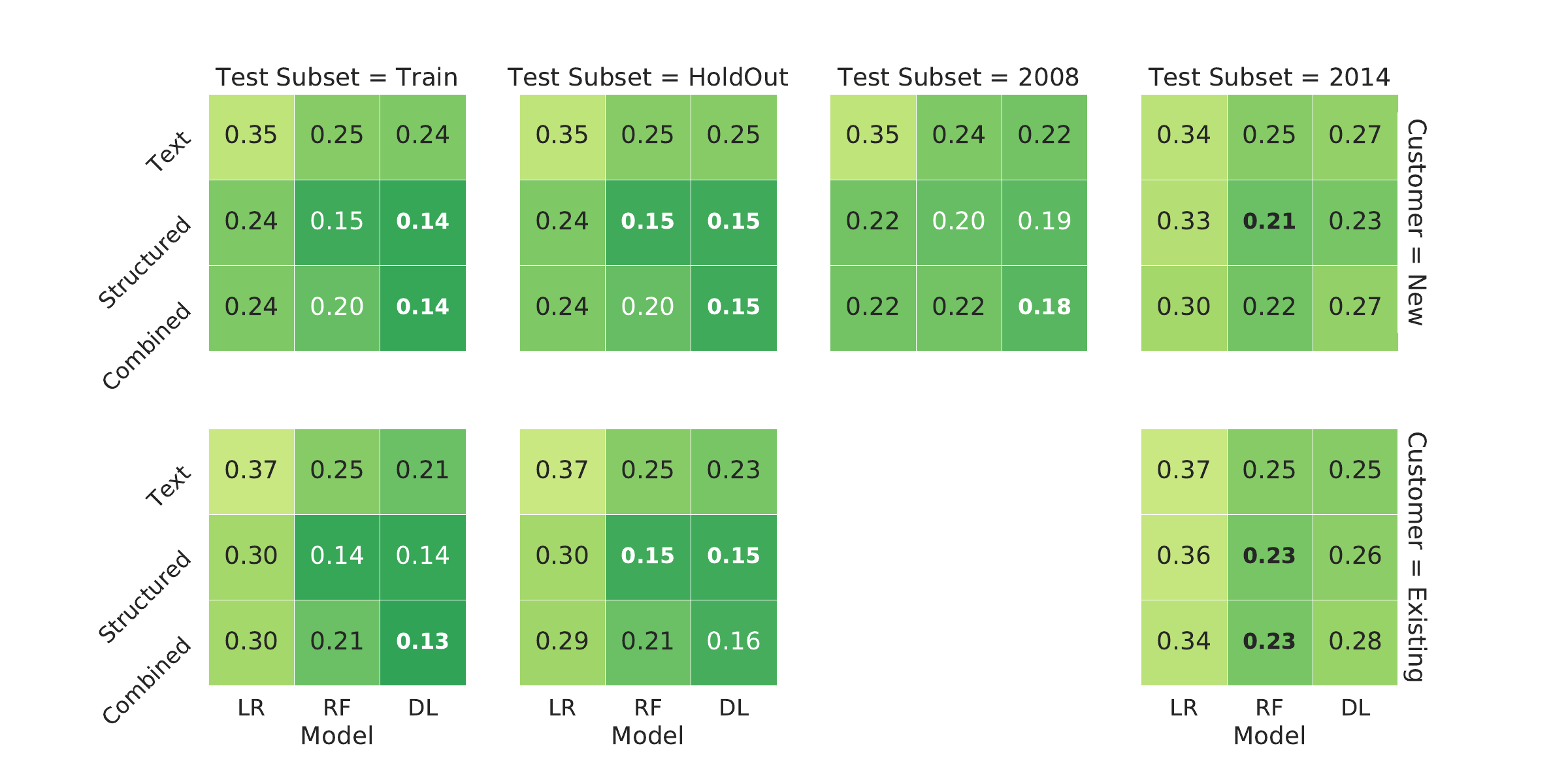}}
  \caption{Brier Model Performance}
  \label{fig:brier_result}
\end{figure}

In addition to the AUC, we also review how well calibrated the model predictions are using the Brier Score metric. The results are presented in Figure \ref{fig:brier_result}. We obtain similar findings as those seen for the AUC results. The non-linear DL and RF models tend to perform better overall compared to the LR across the data and customer subsets, while the DL model outperforms both the LR and RF models in the combined subsets with the exception of the 2014 loan cohort.

From these results, we can conclude that despite the DL text-only BERT model achieving surprisingly high AUC and Brier scores, and the DL model performing well on the structured data, when we combine these inputs there is little to be gained from the introduction of the text. We do find, however, that the DL combined model demonstrates a notable performance uplift over the LR and RF combined models, suggesting that the DL model can better synthesise the multiple inputs. Our findings thus far only consider performance when aggregated over the entire test set. We will explore in section \ref{heading:expimpactoftext} how text, in some instances, may be helpful both for predicting loan default when the structured model is not confident, and demonstrate how this analysis generates new knowledge for business process improvement.

\subsection{The impact of text length on performance} \label{subheading:textlength}

In section \ref{subheading:modelperformance} the results indicated a relative fall in performance in 2014 when the text length was considerably shorter than in the other data subsets. In this section, we explore the role that word count has on performance. As the AUC and Brier scores produced similar findings, we only consider the impact on our primary metric, the AUC score, for conciseness.

\begin{figure}[!htb]
\centering
\scalebox{1}{
  \includegraphics[width=1\linewidth]{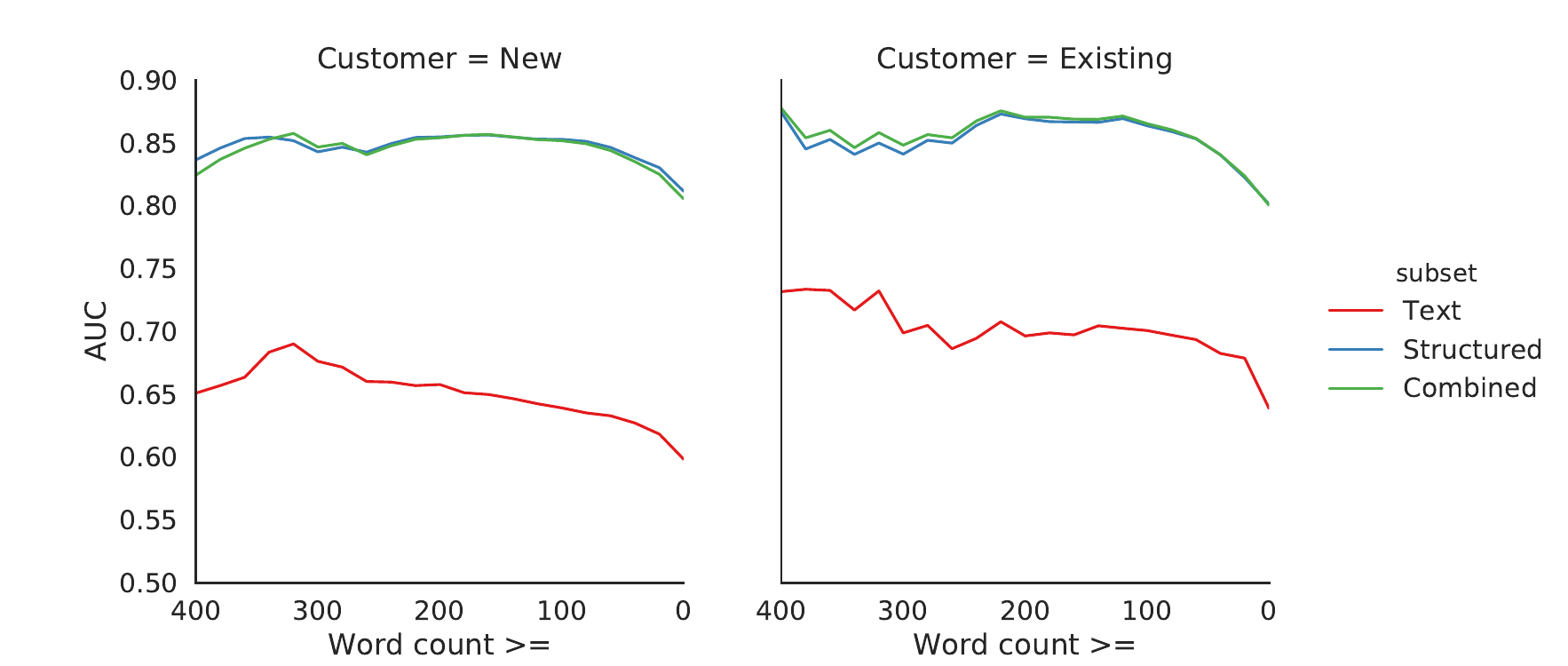}}
  \caption{AUC model performance by word count}
  \label{fig:wc_performance}
\end{figure}

Figure \ref{fig:wc_performance} presents the text word length's cumulative impact on performance. This result is achieved by iteratively calculating the AUC result at different word count intervals across the amalgamated test sets. When the X-axis has reached zero, all cases are included in the result. This figure only presents the DL models' results as they produced the best results for text in the text-only and combined models. The results are presented across all data subsets (Text, Structured, Combined) represented by the line colour, for the two customer types, i.e.\@ new (left) and existing (right) customers. 

Naturally, the text-only models are most sensitive to the input text's length, for both new and existing customers, with longer texts contributing a marked improvement to the AUC results. There appears to be a continuous improvement for the existing customers as word length increases, with some evidence that the improvement tails off beyond 350 words. For the new customer group as well, there is a positive relationship with longer word length; however, beyond 325 words, performance declines again.

Surprisingly, we find that also the structured model performance correlates with the text length. This relationship suggests that the length of the expert's assessment does not vary randomly with the loan's structured characteristics. We hypothesise that certain parts of the input space may have systematically solicited a longer or shorter statement from the loan officer.

The interaction between the structured and combined results is an interesting one. Generally, we can see that the performance across the structured and combined models is broadly similar for both customer groups. For the existing customers, the combined model always performs at least as well as the structured model. Furthermore, beyond 100 words, there is evidence of a slight improvement in the results. This improvement becomes most prominent when the text is over 250 words. The interaction is more complicated for new customers than that seen for existing customers, with extremes in word counts negatively influencing the model performance. 

From a practical perspective, this analysis can help to inform the mSME lending process. For the loan officer who collects the loan, it can be used to provide a target range for word counts. For our data, the target range seems to be around 100 to 325 words, as text statements in this range contribute positively to predictions. Word counts can also be used to measure how much weight to give the text score. For example, for the loan expert who received the loan officer report, text scores with very short or very long word counts should undergo greater scrutiny.

This section concludes that there is an inherent relationship between text length and the text model's predictive performance. Furthermore, when comparing the structured and combined models, performance is similar, suggesting nonetheless that text may lead to marginal improvements in AUC performance within particular word count ranges. In turn, such analysis can be used to inform the mSME loan assessment process. Word count, however, is just one dimension of the text data and is likely a representation of the extent of the detail captured. We will further consider the important words for prediction in section \ref{subheading:importantwords}.

\section{Exploring the impact of text} \label{heading:expimpactoftext}
In the following sections, we further explore what impact adding the text has on the combined model. We first evaluate whether the text is an important feature in the model and how it impacts the structured features' importance. Next, we review the correlation between the predicted probabilities of the structured-only and combined models to see whether the inclusion of text has changed the nature of the predictions. Finally, we focus on the text's content to understand what words are important for cases where the combined model led to improved predictions, and, the structured model was uncertain. 

\subsection{Feature importance analysis} \label{subheading:featureimportance}
In this section, we consider the impact of the text on feature importance.

We evaluate the DL models on the test set for these experiments to obtain the feature importance metrics. Permutation importance is used to measure performance; this method assesses performance by monitoring the change in the relevant metric (in our setting, the AUC) when each feature is randomly permuted. Permutation importance benefits from being compatible with black-box models and is less sensitive to noise than alternative classifier-specific methods such as Mean Gini Decrease \citep{strobl2007bias}. We repeat the feature experiments twice, once with the structured-only data and again with the combined model. The structured DL model feature importance forms the baseline against which we can compare feature importance in the combined model, to see how the relative importance of the structured features changes with the text's introduction. 

\begin{figure}[!htb]
\centering
\scalebox{0.8}{
  \includegraphics[width=1\linewidth]{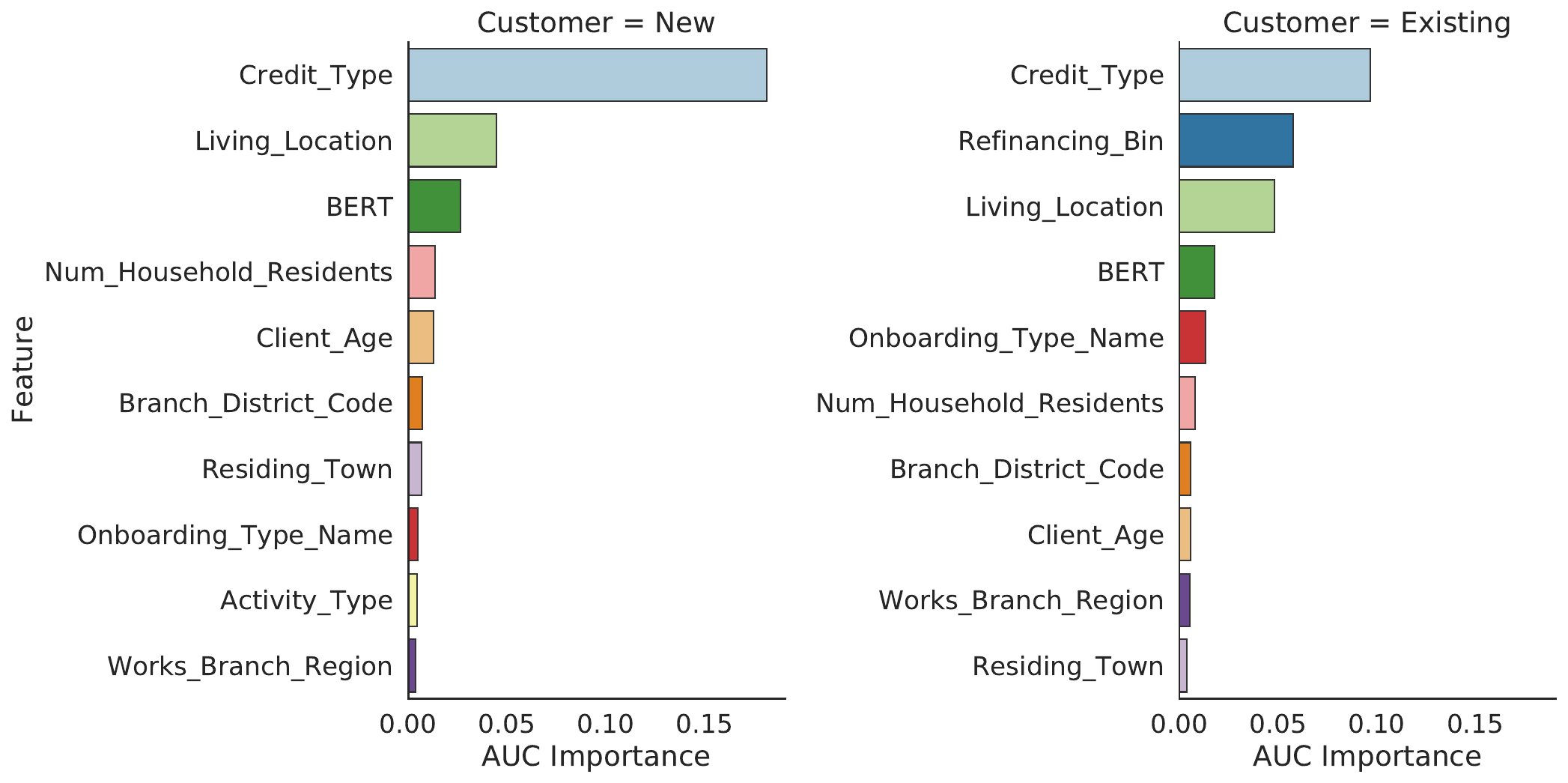}}
  \caption{Feature Importance Ranking of the combined model} \label{fig:VarImp}
\end{figure}

The importance results in the combined model are presented in Figure \ref{fig:VarImp}. Interestingly, we see that the text variable ‘BERT’ is the third most important variable for new customers and the fourth for existing ones. This suggests that the text contains information that is useful for predicting loan default, even with the other variables. Its importance is slightly higher for the new customer group, which is intuitive. The lender has limited information concerning new customers; therefore, additional information from the loan assessment is more useful.

\begin{figure}[!htb]
\centering
\scalebox{1}{
  \includegraphics[width=1\linewidth]{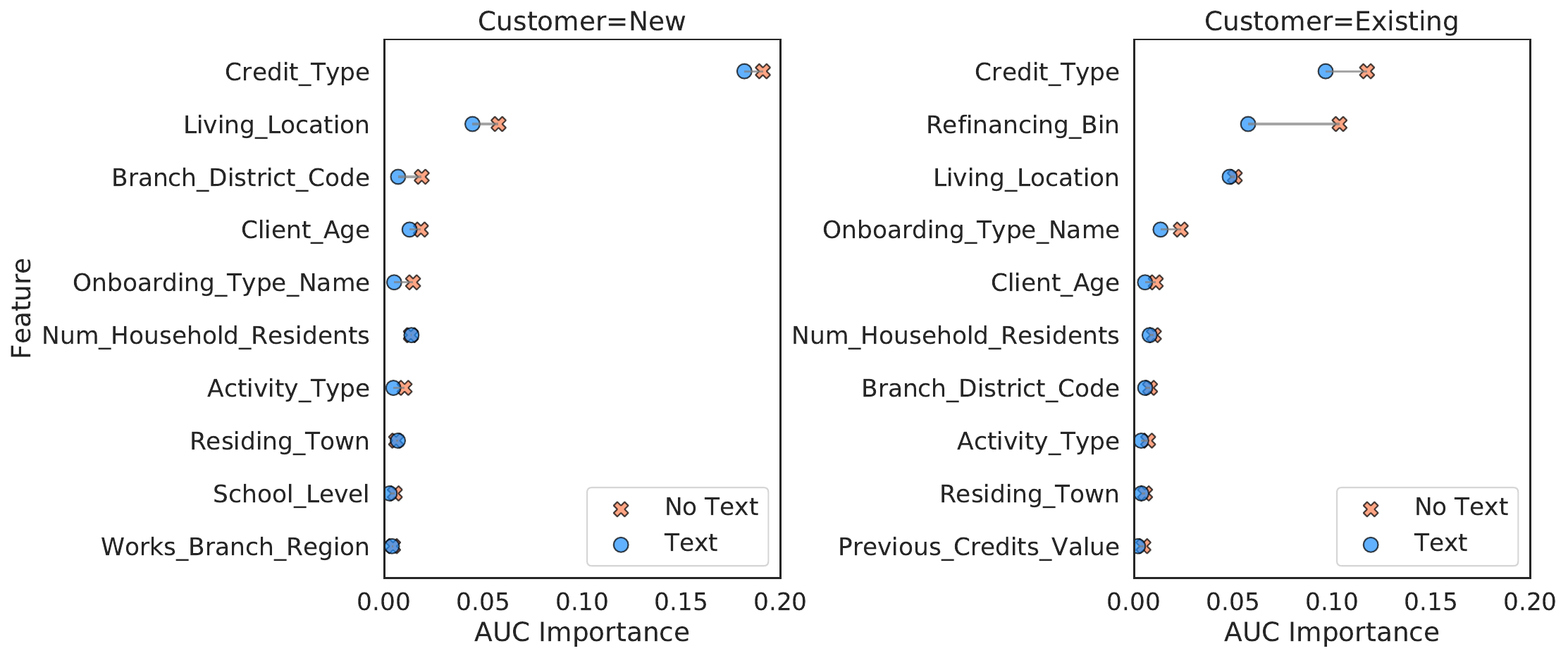}}
  \caption{Change in importance of standard features before/after adding text}
  \label{fig:ImpComp}
\end{figure}

Figure \ref{fig:ImpComp} further shows the extent to which the relative importance of the standard features has changed after introducing the text. The left and right panel shows the shift in importance from the baseline (i.e.\@ the model with standard structured features only) to the combined model's importance.

Across both customer groups, we can see the introduction of text into the model results in a negative shift of importance across many variables. This shift is not surprising as the introduction of an extra variable is, to some extent, expected to make the other variables less important. The more interesting question is which of these variables have seen the most substantial reduction in importance. The most notable reductions in importance for the new customer subset are for Credit Type, Living Location, Branch District Code, while On-boarding Type also becomes somewhat less important. A similar reduction is observed for the existing customer group, although the importance of Living Location is reduced to a lesser extent. The most significant reduction observed across the existing customer group is in the variable signalling refinancing status.

The text's importance and the observed drops in the importance of the structured features suggest that it is not the case that the combined model ignores the text data. Instead, the reduction in the importance of several key features points to why, despite the text alone being predictive, no added performance is gained when it is introduced into the combined model. A plausible explanation for this is that the text contains information that was already partially captured in the structured data (specifically, those variables whose importance is reduced in the combined model). For example, the refinancing status variable could have become less important for existing customers because the applicant's status is also commented on in the text. For this reason, we will further analyse which text words make an actual positive contribution to predictive performance in section \ref{subheading:importantwords}.

\subsection{Impact of text on probabilities } \label{subheading:ImpactonProbabilities } 

To assess to what extent the addition of text in the DL model, and the changes in relative variable importance, actually affect the probability estimates of being in default, we proceed by reviewing the correlation between the different sets of model outputs. 

\begin{table}[!htb]
\caption{Rank order correlation between default probability estimates (test set)}
\centering
\scalebox{0.8}{
\begin{tabular}{llccc}
\hline
Customer           & Data Subset    & Structured & Text & Combined Model \\ \hline
New Customers      & Structured     & 1.00       &      &                \\
                   & Text           & 0.21       & 1.00 &                \\
                   & Combined Model & 0.97       & 0.36 & 1.00           \\
                   &                &            &      &                \\
Existing Customers & Structured     & 1.00       &      &                \\
                   & Text           & 0.19       & 1.00 &                \\
                   & Combined Model & 0.96       & 0.34 & 1.00           \\ \hline
\end{tabular}}
\label{CorrTable}
\end{table}

In Figure \ref{CorrTable}, we opt for the Spearman Rank definition of correlation as it is non-parametric and allows us to gain insight into whether the ordering of the probabilities (and, hence, the risk ranking of applicants) has fundamentally changed in the aggregated test sets. There appears to be little agreement between the text-only model and the structured model as they are weakly correlated, with correlation coefficients of 0.19 and 0.21 for new and existing customers, respectively. Furthermore, across both subsets of customers, the combined model is highly correlated with the structured data, showing correlation coefficients greater than 0.95. High correlation coefficients between the combined and structured models suggest that the structured data has dominated the model, as we might expect given the richness of the structured features and the similarity of the performance results reported in Section \ref{heading:results}.

\subsection{Analysis of text content using LIME} \label{subheading:importantwords}
The feature importance analysis has demonstrated how the model has, to some extent, been influenced by the text, while the correlation analysis revealed a modest shift in the ranking of the probability outputs on the test set. This section reviews under which circumstances the text's content leads to improved performance of the combined model.

To assess which words lead to an improved result, we review cases where the structured-only model prediction of default was uncertain, and an improvement was observed in the combined model prediction. These are, therefore, instances in which the inclusion of text had introduced additional value. For this analysis, we consider uncertain cases as those where the structured-only model produced predicted probabilities between 0.40 and 0.60. Of this subset, we review the 250 cases where the greatest improvement was seen in the test set for the combined model.

To examine these 250 cases, we apply LIME \citep{ribeiro2016should}, an approach that can interpret individual examples classified by a black-box model. In essence, LIME assumes that while the overall relationships between inputs and output may be non-linear, at a local level, linear surrogate models can be trained to observe how an individual prediction changes when the data is perturbed. In the context of textual analysis, the approach perturbs examples by iteratively removing words from the example. We use LIME in favour of alternative interpretive methods such as SHAP \citep{NIPS2017_7062} due to its relative computational efficiency given the large number of parameters in the DL model. An alternative approach to exploring text importance might also be to review the attention maps of the BERT transformer model; however, unlike LIME, this would solely provide the magnitude while ignoring the direction of the output prediction which we seek to understand.

\begin{figure}[!htb]
\centering
\scalebox{0.6}{
  \includegraphics[width=1\linewidth]{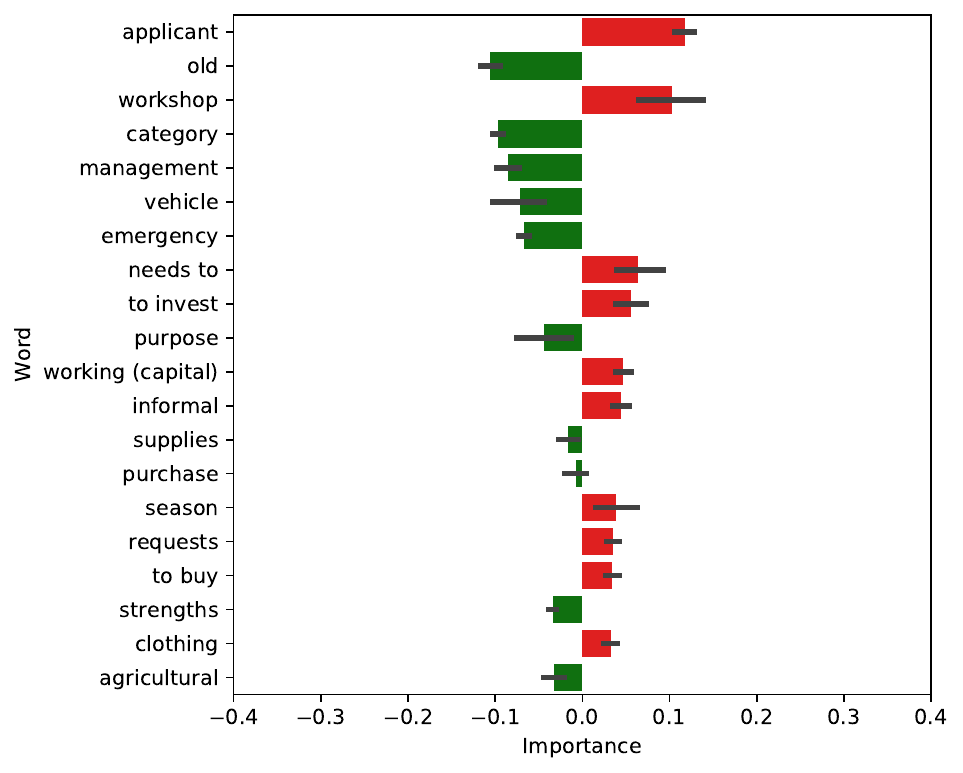}}
  \caption{Top 20 important Words for cases where text led to improved predictions of default}
  \label{fig:LIME_Top_250}
\end{figure}

The twenty most important words across this set of statements are presented in Figure \ref{fig:LIME_Top_250}. The bar's length represents the mean impact on the output prediction and, therefore, the overall importance. Positive values indicate that the word's inclusion increased the predicted probability of default and, conversely, negative values demonstrate a reduced predicted probability of default. To arrive at the final subset of 20 words, we apply post-processing for greater interpretability, including word stemming and the removal of stop words. 

From Figure \ref{fig:LIME_Top_250}, several themes can be identified relating to the nature of the organisation, the loan requirement and the semantics of the words the loan officer chooses.

The occurrence of some terms suggests that the model has placed importance on industry type, with words such as `workshop', `clothing' and `agricultural'. This is notable as the model has placed importance on a more granular description of the industry than the structured feature `industry type' provides. Interestingly, the model associated a low risk to `agriculture' and a higher risk to the words `workshop' and `clothing' businesses. The term `informal' is also used to describe some applicants and leads to increased predictions of default. In this context, the word `informal' describes an organisation that is not tax/VAT registered reflecting a smaller organisation with lower income.

Other important words include those which reference the nature of the loan requirement. Words such as `supplies', `vehicle' and `purchase' suggest an asset driven investment and a lower predicted probability of default. Furthermore, the word `emergency' suggests a short term loan requirement to resolve a temporary cash-flow issue and has a lower associated default prediction. Conversely, the word `working (capital)' may suggest a longer-term credit requirement and an increased default prediction. The word `season' also leads to an increased predicted likelihood of default and tends to refer to a temporal investment, for example; a market seller may look for a loan to invest for the festive `season' and likewise a farmer might look for credit to fund the growing `season'.

While elements such as the nature of the loan and the industry the organisation operates within are factually based, there is arguably some evidence to suggest the model has also learned the implicit semantics of the loan officer's text statement. For example, two of the most important words are `applicant' and `old (customer)'. Both are used to describe the customer; however, the term `old' customer alludes to an inherent trust from the loan officer, which, in these instances, is useful for prediction. An alternative theory might be that this is measuring the same information provided in the structured feature `Credit Type'. This would align with the conclusions in section \ref{subheading:featureimportance} where the variable `Credit Type' saw the greatest reduction of importance when the text was introduced. The word `strengths' is also used in some statements which is an explicit endorsement of the applicant and results in lower model predictions of default. Furthermore, this also suggests that loan officers have a good ability to determine the risk of good borrowers. The way the loan requirement is framed also appears to impact the output prediction. For example, while `purchase' reduces the prediction of default, `to invest', `needs to', `requests' and `to buy' lead to higher predicted probabilities. In these instances, the words themselves provide little factual information yet are important for prediction, suggesting the models' interpretation of these words is implicit.

\begin{figure}[!htb]
\centering
\scalebox{1}{
  \includegraphics[width=1\linewidth]{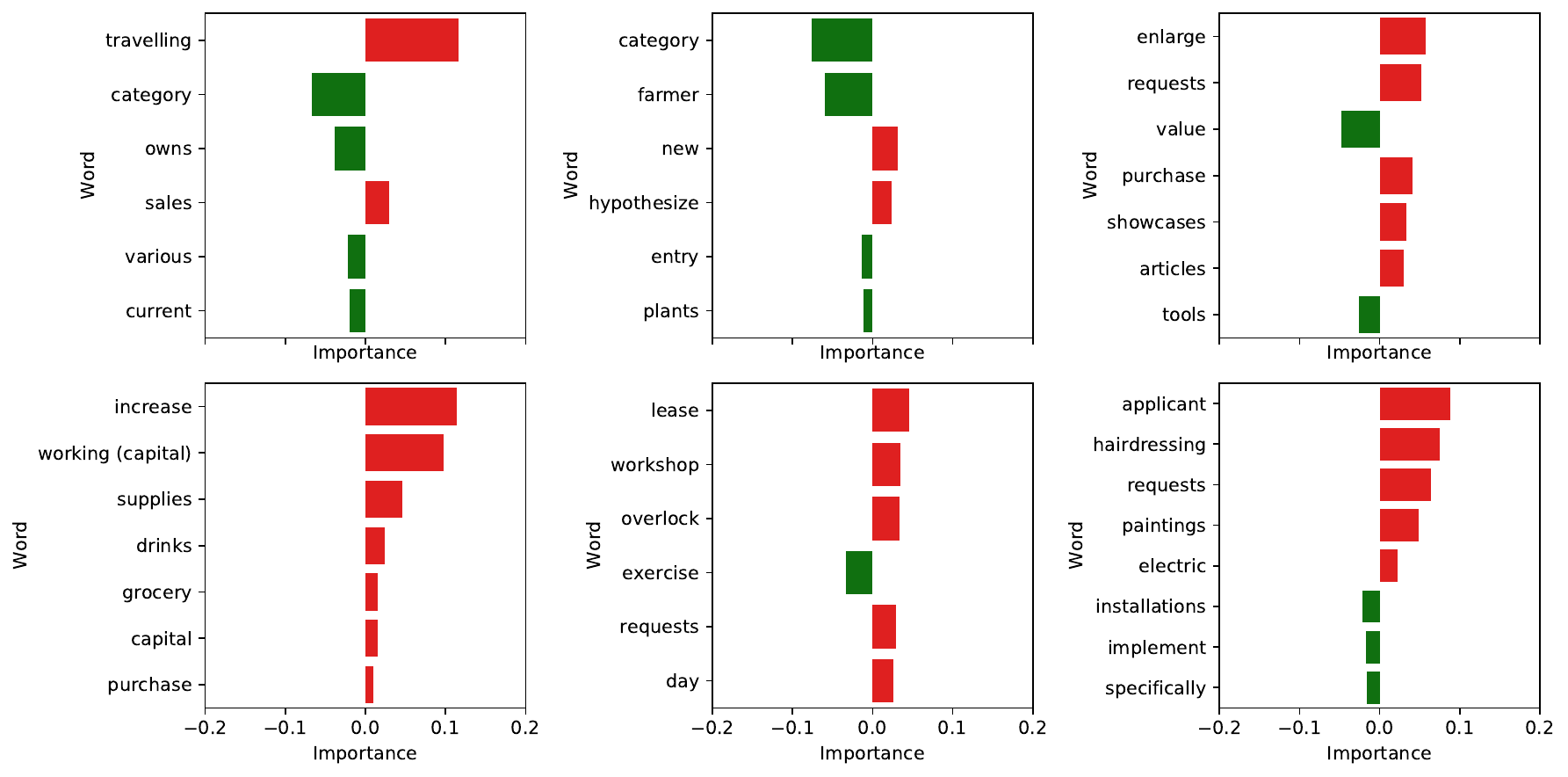}}
  \caption{Examples of word importances for individual cases}
  \label{fig:LIME_IndiExamples}
\end{figure}

Figure \ref{fig:LIME_IndiExamples} shows the importance for six individual cases where each tile represents a unique case sampled from the top 250. This helps to understand what the model has learned at a more granular level. Furthermore, from an operational perspective, such tools can build trust in the models and help them identify further opportunities to improve future loan assessments' quality. The plot confirms the findings of the aggregated top 20 words in Figure \ref{fig:LIME_Top_250} with the presence of themes surrounding the nature of the organisation, the loan requirement and the semantics of the text. Notably, this more granular view reveals some differences between the industry types that were not included in the top 20 with the word `farmer', or `agricultural', appearing to reduce predicted risk. Furthermore, the words `travelling (market stall)', `hairdressing' and `grocery' produce higher risk scores.

This analysis of the text content using LIME demonstrates how text can provide value for mSME lenders. Although our predictive results suggest that, on aggregate, the text does not introduce additional value for our dataset, it is not necessarily the case that the text is never useful as we find some reoccurring themes that do add potential value. From a management perspective, identifying these themes can help shape future loan officer assessments by knowing which topics to focus on in the written statements. We speculate that this may indeed lead to improved performance over using the structured features alone. The other exciting application for the output of such content analysis is to `structure the unstructured', capturing key themes such as the loan requirement, tax status and more granular industry types as standard model features.

\section{Summary and implications}\label{heading:summaryandimplications}

The first aim of this paper was to explore whether the statements produced by the loan officer are predictive of loan default, and, whether Deep Learning provides improved performance over traditional text mining and machine learning approaches. We benchmarked the DL BERT model's performance against an LR and RF model that uses LSA to extract features from the text. Our results suggest that text can be surprisingly predictive of loan default, producing AUC scores between 0.55 and 0.71. Furthermore, we find that the DL BERT model provides a material uplift over the baseline text models ranging between 7\% and 25\%. The only exception to these findings is observed in the 2014 out-of-time sample, for new customers, when all models performed poorly due to both a reduction in text length and population drift.

The second aim was to test whether including the loan officer's text produces better predictions of mSME loan default than using the structured data alone. Furthermore, we sought to understand whether a Deep Learning approach could better synthesise the combined structured and text inputs. Our findings relating to this are mixed. Despite promising results from the DL BERT model on the text alone, when we review the performance of the combined models which use both the text and structured features as inputs, we find limited evidence that the contribution of the text would lead to improved performance --- both in terms of discriminative capacity (AUC) or how well-calibrated the models are (Brier Score). While the RF model performance deteriorated with the introduction of text features, the LR and DL models maintained the performance seen for the structured models, with the DL model, in general, being the best performing. We further explored this phenomenon by reviewing the AUC performance across word count thresholds, which we suspected might have influenced the results given the 2014 out-of-time sample's poor performance. Intuitively, we find some evidence that longer statements improve the text-only DL BERT model performance, with statements between 100 and 325 words long leading to the slight improvement of the combined models over the structured-only models. Therefore, we argue that textual data for mSME credit scoring holds clear potential, but unsurprisingly, a positive result depends on the quality of the text. Nonetheless, the DL combined model performance seemed to be reasonably robust to shorter text; that is to say, these statements did not significantly deteriorate the model performance. In this respect, we argue that the DL combined model may be suitable for supporting the credit expert's role by providing them with integrated risk scores. Furthermore, we may reasonably expect that, for other less developed datasets with fewer structured features, the text can produce an added performance lift given the predictive power of the text-only model. It may also mitigate the requirement for acquiring costly external variables.

Our third aim was to explore how text information changes the model behaviour and associated predicted probabilities. To address this question, we reviewed the text's variable importance and compared the relative change in the importance of the structured features with and without the text being present. From the variable importance analysis, we find that the text variable `BERT' is important, as it features in the top four most important variables across both subsets of customers. The text's importance is significant since it shows that, although the text neither improves nor harms model performance, it is not the case that the combined DL model simply ignores the text features. Furthermore, when we review the relative importance of the structured variables, we observe that some structured variables fall in importance when the text is added to the model. Again, the shift in importance shows that the text has changed the behaviour of the model. We argue that this drop in importance can be attributed to the duplication of information between the assessment text and the structured features. This duplication of information may partially explain why the text does not improve the performance metrics. To understand to what extent the addition of text affects the risk ranking, we inspected the Spearman rank correlation. A high coefficient of over 0.95 between the structured-only and combined DL model predictions confirms that the combined model outputs are more strongly impacted by the structured data.

This paper's final aim was to explore how the actual text content influences the predictive capacity of the text. To address this, we conducted a LIME analysis looking at the important words featured in cases where the structured-only model was uncertain what to predict and adding the text input led to a better prediction. We thus identify several reoccurring themes to which the text model appears to assign importance. These include information relating to the nature of the organisation, the loan requirement, and the sentiment projected by the loan officer. This analysis highlights how the lender could improve future reporting by its loan officers to achieve better predictions of loan default. Furthermore, aspects of the identified topics such as granular industry types and the loan requirement (e.g.\@ working vs emergency loan) may, alternatively, be captured under the form of structured data. This highlights a useful secondary use of the text which a lender may look to repeat on an ongoing basis.

\section{Conclusions and further research}\label{heading:conclusion}
In summary, our research has demonstrated that deep learning approaches can be useful for mSME default prediction, as, on most of our test samples, a novel deep learning approach using the BERT model outperformed two benchmark models -- Logistic Regression and Random Forests. Although our performance results suggested that textual loan information does hold predictive capability, producing fairly accurate predictions when used on its own, this did not lead to an aggregate performance lift when added to a model that already had the structured features included. Further analysis, however, shows that appropriately sized texts may provide a marginal performance improvement over the structured data alone. We have also shown how an exploration of the text can help identify which new structured features to collect or what aspects to focus on in the expert loan assessment. For future research, we believe it would be interesting to explore how, on alternative mSME lending datasets, the textual loan assessments impact performance, as datasets with insufficient structured data may benefit more from the inclusion of text. Of course, the fundamental structure of the text from other lenders will undoubtedly influence the result. We reviewed a subset of examples using LIME to see which words in the text are associated with improved predictions, and, from that, were able to formulate practical recommendations for writing mSME loan text assessments. Though out of the scope of this paper, for further research, a follow-up analysis showing what topics are important for different customer types has the potential to produce further recommendations. 

\section*{Acknowledgements}
This work was supported by the Economic and Social Research Council (grant nr. ES/P000673/1). The last author acknowledges the support of the Natural Sciences and Engineering Research Council of Canada (NSERC) (Discovery Grant RGPIN-2020-07114). This research was undertaken, in part, thanks to funding from the Canada Research Chairs program. The authors acknowledge the use of the IRIDIS High Performance Computing Facility, and associated support services at the University of Southampton, in the completion of this work.


\end{document}